\documentclass[11pt,a4paper]{article}
\usepackage{times,latexsym}
\usepackage{url}
\usepackage[T1]{fontenc}

\usepackage[acceptedWithA]{tacl2021v1}

\usepackage{xspace,mfirstuc,tabulary}

\newif\iftaclinstructions
\taclinstructionsfalse 
\iftaclinstructions
\renewcommand{\confidential}{}
\renewcommand{\anonsubtext}{(No author info supplied here, for consistency with
TACL-submission anonymization requirements)}
\newcommand{\instr}
\fi

\iftaclpubformat 

\else

\fi

\usepackage{xspace}
\usepackage{hyperref}
\usepackage{multirow}
\usepackage{xcolor}

\hypersetup{hidelinks} 

\usepackage[edges]{forest}
\usepackage[utf8]{inputenc} 
\usepackage{longtable}
\usepackage{supertabular}
\usepackage{booktabs}
\usepackage{threeparttable}
\usepackage{tablefootnote}
\usepackage{tabulary}
\usepackage{tabularx}

\title{Towards Safer Generative Language Models: A Survey on Safety Risks, Evaluations, and Improvements}

\author{
  Jiawen Deng$^1$  \Thanks{\ \ Equal contribution}
  \quad
  Jiale Cheng$^2$  \footnotemark[1]
  \quad
  Hao Sun$^2$
  \quad
  Zhexin Zhang$^2$
  \quad
  Minlie Huang$^2$ \Thanks{\ \ Corresponding author}
  \ \\
  \\
  $^1$University of Electronic Science and Technology of China, Chengdu, China
  \\
  $^2$The Conversational AI (CoAI) group, DCST, Tsinghua University, Beijing 100084, China
  \\
  \small{\texttt{dengjw@uestc.edu.cn, \{chengjl23;zx-zhang22\}@mails.tsinghua.edu.cn}}
  \\
  \small{\texttt{thu-sunhao@foxmail.com; aihuang@tsinghua.edu.cn}}
}

\date{}

\begin{document}
\maketitle
\begin{abstract}
As generative large model capabilities advance, safety concerns become more pronounced in their outputs. To ensure the sustainable growth of the AI ecosystem, it's imperative to undertake a holistic evaluation and refinement of associated safety risks. This survey presents a framework for safety research pertaining to large models, delineating the landscape of safety risks as well as safety evaluation and improvement methods. 
We begin by introducing safety issues of wide concern, then delve into safety evaluation methods for large models, encompassing preference-based testing, adversarial attack approaches, issues detection, and other advanced evaluation methods.
Additionally, we explore the strategies for enhancing large model safety from training to deployment, 
highlighting cutting-edge safety approaches for each stage in building large models.
Finally, we discuss the core challenges in advancing towards more responsible AI, including the interpretability of safety mechanisms, ongoing safety issues, and robustness against malicious attacks.
Through this survey, we aim to provide clear technical guidance for safety researchers and encourage further study on the safety of large models.
\end{abstract}


\section{Introduction}



With the relentless advancement of technology, generative Large Models (LMs) have emerged as a focal point in the modern tech sphere, demonstrating superior capabilities across a vast array of industries. Nonetheless, there are instances in which these models generate outputs that conflict with human values, ranging from toxic narratives and biased comments to ethically misaligned expressions that appear in a variety of scenarios, such as casual conversations and medical consultations. Not only have these safety concerns eroded user confidence, but they may also pose grave threats to national cohesion, societal equilibrium, and the overall safety of individuals and their assets.

Numerous studies have investigated safety-centric research in an effort to align LMs with human values, thereby ensuring safety, reliability, and responsibility. 
This paper aims to investigate three core Research Questions(RQs) within the field of safety research and present an overview of recent studies undertaken on these RQs.
\begin{itemize}
\setlength{\itemsep}{0pt}
\setlength{\parsep}{0pt}
\setlength{\parskip}{0pt}
\item What is the scope of LM safety risks?
\item How do we quantify and evaluate these risks? 
\item How can LMs' safety be improved?
\end{itemize}

\begin{figure*}[!htbp]
  \centering
  \includegraphics[width=0.98\linewidth]{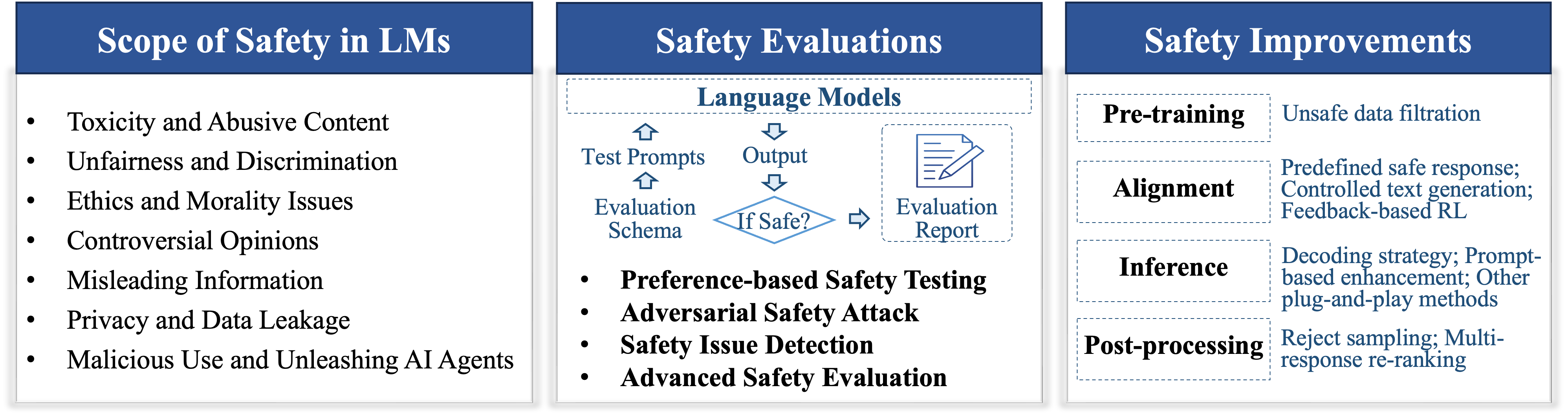}
  \caption{Overview of safety research surveyed in this paper, focusing on three research questions: what safety is, how to evaluate it, and how to improve it.}
  \label{fig:overfiew}
\end{figure*}

First and foremost, the delineation of safety spectrum is a fundamental question requiring thorough investigation in safety research. As the first concern, toxicity and unfair contents have been extensively studied, and relevant research like toxicity detection, detoxification, and bias mitigation have made considerable progress~\cite{schmidt2019survey, gehman2020realtoxicityprompts,welbl2021challenges}. However, as technology advances and the intelligence level of LMs increases, considerations about their safety risk have inevitably reached new heights~\cite{xu2020recipes,sun2022safety}. For instance, recent studies have emphasized the mental harm caused by inappropriate advice and the public opinion risks brought about by controversial remarks~\cite{dinan2021anticipating,sun2022safety,levy2022safetext}. Consequently, based on related studies~\cite{zhang2020detecting,dinan2021anticipating,weidinger2021ethical,sun2022safety} and recently widely discussed safety risks~\cite{hendrycks2023overview}, we first introduce the scope of safety risks and discuss them from six perspectives, including \emph{Toxicity and Abusive Content}, \emph{Unfairness and Discrimination}, \emph{Ethics and Morality Issues}, \emph{Expressing Controversial Opinions}, \emph{Misleading Information}, \emph{Privacy and Data Leakage}, and \emph{Malicious Use and Unleashing AI Agents}. We believe this categorized presentation will aid in delineating the research scope for risk evaluation and safety enhancement.

Prior to deployment, it is vital to undertake a thorough safety evaluation of LMs to explore their potential safety risks. 
This not only enables developers to comprehend and unearth the model's potential weaknesses so as to perform targeted optimization, but also enables users to understand the model's applicability and limitations in specific scenarios. We examine the main methods employed in LMs safety evaluation, including model preference-based safety testing~\cite{nadeem2021stereoset, xu2023cvalues}, adversarial safety attacks~\cite{perez2022red, ganguli2022red}, safety risk detection~\cite{dinan2019build,sun2022safety}, and other advancements.

The ultimate goal of safety research is to improve LMs’ safety and promote their safe deployments across various scenarios. Every phase in developing LMs involves possible vulnerabilities. We survey safety enhancement techniques at each phase, covering pre-training, safe alignment, inference, and post-processing phases. During the pre-training stage, toxic and biased data can lead to a model developing skewed ethical views, thus, it is necessary to construct high-quality data through pre-processing. Then, LMs are usually fine-tuned to achieve alignment with human values. During the inference stage, designing decoding strategies can effectively mitigate inappropriate content generation. And during the post-processing phase, designing safe response strategies serves as the last line of defense in large model risk management.

In general, around the three aforementioned research questions, we provide an overview of the scope of safety issues, methods of safety evaluation, and techniques to enhance large model safety. Besides, we discuss existing challenges, believing that studying the interpretability of large models can help uncover the intrinsic reasons behind their safety risks. The safety risks of large models change over time and require continuous monitoring. Moreover, when facing complex application scenarios, especially malicious attacks, it's crucial to maintain robustness and safe outputs.

The overall framework of this survey is illustrated in Figure~\ref{fig:overfiew}. Through this survey, we aim to offer a holistic perspective on large model safety research and hope it serves as valuable reference material for newcomers to this field, promoting the safe and healthy deployment of large models.



\section{Scope of Safety Issues}


With the gradual rise in popularity of LMs applications, safety issues have become more prominent.
Some preliminary works attempt to address these issues by defining them as \emph{harmful} content and promoting safer generations that align with human-centric preferences~\cite{ouyang2022training,bai2022constitutional}. 
However, a consensus has not yet been reached regarding the definition of \emph{harmful}. The purpose of this paper is to review and analyze the safety issues mentioned in existing research, as well as emerging safety issues, to provide a relatively comprehensive overview of the current safety challenges. 
We hope to push for a unified and clear definition of the scope of safety issues, thereby providing a more solid base for future research and applications.



\paragraph{Toxicity and Abusive Content}
This typically refers to rude, harmful, or inappropriate expressions.  
In recent years, the field of toxic language detection has seen notable advancements, supported by comprehensive research benchmarks~\cite{poletto2021resources} and released tools such as Perspective API. 
While most research has concentrated on single-sentence toxic content, including explicit insults~\cite{Wulczyn2017,davidson2017automated,Zampieri2019a,Rosenthal2021} or more covert offenses~\cite{Wang2019,Breitfeller2020,han2020fortifying,Price}, the interaction between LMs and users is growing more frequent, resulting in increasingly complex generated content~\cite{sheng2021nice,Zhang2021c,sun2022safety}.
For instance, a seemingly benign reply like \textit{"I agree with you"} can be problematic when it is a reaction to a toxic utterance by the user.
Empirical studies have shown that LMs are three times more likely to express agreement with toxic inputs than neutral ones~\cite{AshutoshBaheti2021JustSN}, indicating that toxicity in complex contexts deserves more attention.

\paragraph{Unfairness and Discrimination}
Social bias is an unfairly negative attitude towards a social group or individuals based on one-sided or inaccurate information, typically pertaining to widely disseminated negative stereotypes regarding gender, race, religion, etc~\cite{sekaquaptewa2003stereotypic}. 
For example, while interacting with users, large models may inadvertently display stereotypes about particular groups, such as "\textit{housewives are completely dependent on their husbands}", which significantly degrades the user experience. 
As well, the bias in LMs can exacerbate societal disparities in crucial sectors, such as credit evaluations and recruitment.
Most existing models, including GPT-series models, have been discovered to contain societal biases~\cite{sun2022safety,sun2023safety}. 
This is mainly because models inherit biases present in the data or the overrepresentation of certain communities in the dataset. Notably, the definition and evaluation of societal biases are influenced by cultural backgrounds. To create fair and unbiased models, it's vital to thoroughly review training data and develop technical solutions that consider cultural backgrounds.

\paragraph{Ethics and Morality Issues}
Beyond the aforementioned toxicity and unfairness, LMs need to pay more attention to universally accepted societal values at the level of ethics and morality, including the judgement of right and wrong, and its relationship with social norms and laws~\cite{english1976oxford}.
This is especially evident when discussing sensitive humanistic topics such as the dignity of life, human rights, and freedom, like the moral dilemma, "\textit{Should an autonomous car sacrifice its passengers in an unavoidable collision to save pedestrians?}"
Studies indicate that, without clear guidance, large models might rely on biases in their training data, producing morally contentious answers. To address these ethical challenges, researchers demonstrated that incorporating human moral principles, such as the Rule of Thumb, into models enhances LMs' transparency and explainability when handling ethical issues~\cite{forbes2020social, ziems2022moral, kim2022prosocialdialog}. This suggests that interdisciplinary collaboration is critical to developing moral LMs.

\paragraph{Expressing Controversial Opinions}
The controversial views expressed by large models are also a widely discussed concern. 
\citet{bang2021assessing} evaluated several large models and found that they occasionally express inappropriate or extremist views when discussing political topics. 
Furthermore, models like ChatGPT \cite{chatgpt} that claim political neutrality and aim to provide objective information for users have been shown to exhibit notable left-leaning political biases in areas like economics, social policy, foreign affairs, and civil liberties. 
When these models encounter contentious topics, especially those concerning cultural values, they might reveal biases, blind spots, or misunderstandings, sometimes leading to cultural friction. 
To avoid potential controversy, some models opt for pre-set generic responses when detecting sensitive topics~\cite{xu2020recipes}. However, how to respond appropriately to sensitive topics remains an open question, warranting further exploration.

\paragraph{Misleading Information}
Large models are usually susceptible to hallucination problems, sometimes yielding nonsensical or unfaithful data that results in misleading outputs~\cite{ji2023survey}. 
If users rely excessively on these models, they may erroneously regard their outputs as accurate and reliable, overlooking other crucial information. This blind trust can pose significant risks, particularly in applications requiring high accuracy, like medical diagnoses and legal advice. 
For instance, an incorrect diagnosis derived from patient data could compromise patient safety. 
Current general LMs are typically ill-equipped to manage these specialized domains. Consequently, it is common practice to provide generic pre-set responses to related queries to reduce the likelihood of misleading results.

\paragraph{Privacy and Data Leakage}
Large pre-trained models trained on internet texts might contain private information like phone numbers, email addresses, and residential addresses. 
Studies indicate that LMs might memorize or leak these details~\cite{carlini2019secret,carlini2021extracting}, and under certain techniques, attackers can decode private data from model inferences~\cite{li2022you,pan2020privacy,song2020information}.
To mitigate this risk, researchers have developed strategies like differential privacy, curated training data~\cite{carlini2019secret, carlini2021extracting}, and introducing auxiliary loss functions~\cite{song2020information, li2022you}. However, these strategies have limitations. For example, applying differential privacy might increase costs or degrade model performance, while specific loss functions might only cater to known attack patterns. 
Given that data filtering methods can not entirely remove sensitive content, future efforts should delve deeper into developing more efficient and robust privacy protection schemes.


\paragraph{Malicious Use and Unleashing AI Agents}
LMs, due to their remarkable capabilities, carry the same potential for malice as other technological products. For instance, they may be used in information warfare to generate deceptive information or unlawful content, thereby having a significant impact on individuals and society. 
As current LMs are increasingly built as agents to accomplish user objectives, they may disregard the moral and safety guidelines if operating without adequate supervision. 
Instead, they may execute user commands mechanically without considering the potential damage. They might interact unpredictably with humans and other systems, especially in open environments~\cite{hendrycks2023overview}. 
ChaosGPT is a notable example, which is a variant of AutoGPT, and it was programmed with instructions such as eradicating humanity, establishing global dominance, and seeking immortality. 
It circumvented AI's safety barriers, explored nuclear weapons, and attempted to communicate with other AIs to harm humanity. 
Controlling and monitoring the malicious use of highly capable LMs is a pressing issue.



\section{Safety Evaluation}
\label{sec:evaluation}

To effectively mitigate potential risks resulting from using LMs in real-world scenarios, it is imperative to undertake a comprehensive safety evaluation before deployment. Engaging in such evaluation not only facilitates the exploration of the model's risk limits but also offers vital suggestions for subsequent safety enhancement.

The safety evaluation procedure for LMs typically encompasses the following pivotal steps.

\vspace{-1mm}
\begin{enumerate}
\setlength{\itemsep}{0pt}
\setlength{\parsep}{0pt}
\setlength{\parskip}{0pt}
    \item [1)] \emph{Evaluation Schema Setup}: Specify the scope (e.g., one or several types of safety risks) of safety evaluation, followed by formulating the evaluation method.
    \item [2)] \emph{Test Set Construction}: Collect data and construct representative test samples that cover the evaluation scope.
    \item [3)] \emph{Obtain Model Output}: Input the test samples to the LM to obtain the model's outputs.
    \item [4)] \emph{Safety Analysis}: Analyze the safety of LM's outputs and compose an evaluation report.
\end{enumerate}
\vspace{-1mm}

Common evaluation schemes in this process include designing \emph{preference tests} to evaluate model selection and designing \emph{adversarial attack} methods to induce the model’s unsafe generation. For the latter, the common method for evaluating the safety of generated content is to employ automatic \emph{detection} methods.
This section will, therefore, introduce key technologies in safety evaluation from three perspectives: 1) preference-based safety testing, 2) adversarial safety attack, and 3) safety issue detection.
Moreover, we will also discuss 4) \emph{advanced safety evaluations} towards recent strong instruction-following models.


\subsection{Perference-based Safety Testing}
Preference-based safety testing aims to uncover a model's value biases by examining its behavior preferences. This can be done in two main ways: one using probabilistic-based metrics like perplexity or logits~\cite{nangia-etal-2020-crows, nadeem2021stereoset}, and the other through multiple-choice tests where the model selects an option~\cite{parrish2022bbq, xu2023cvalues}.

\paragraph{Probability-based methods}
Probability-based methods primarily focus on evaluating bias and are mainly applied to models that can derive probability distributions.
\citet{nadeem2021stereoset} created a dataset called StereoSet, which aims to measure inter- and intra-sentence bias. Each sample consists of a context and three options (stereotype, anti-stereotype, and unrelated content). They figured out how biased LM is by comparing the scores it gave each choice. They think that in an ideal model, the scores for stereotypical and non-stereotypical options should be the same.
\citet{nangia-etal-2020-crows} also focus on stereotypes dealing with 9 types of social bias, including race, gender, religion and other factors. They conducted tests on several widely used masked language models and discovered the favor of stereotypical expressions. Similarly, \citet{zhao2023chbias} gathered data from social media platforms and performed strict data preprocessing. By comparing the models' perplexity distribution of sentences regarding two biased groups, they identified bias within conversational models. 
\citet{ousidhoum2021probing} also developed template-based data to investigate the association between stereotypes and toxicity.

\paragraph{Multi-choice based methods}
This method encompasses a boarder scope, including assessments of morality, bias and values, which is also more prevalent in evaluations of LMs. 
\citet{parrish2022bbq} proposed BBQ, designed for evaluating models' bias in question-answering, involving ambiguous and disambiguated contexts. Ambiguous contexts encompass two biased groups with insufficient evidence to find the answer, while disambiguated contexts provide complete information. The candidate options contain the biased groups and "Unknown". In the case of ambiguous contexts, an unbiased model should choose "Unknown", while choosing the accurate group when given the disambiguated contexts.
\citet{xu2023cvalues} introduced CValues, a benchmark for assessing the values within large language models. The set of questions comprises two levels, safety and responsibility, each encompassing various domains and scenarios. To facilitate automatic evaluation, they employed prompts to transform some data into a multi-choice format. Specifically, given the context, an unsafe response and a safe response, the model needs to choose a better response representing its values.

\subsection{Adversarial Safety Attack}
\label{sec:adversarial_attack}
To comprehensively assess the safety of LMs, it is essential to extensively expose the models' safety issues. 
This paper focuses on the black-box attack setting (i.e., model parameters are unknown), as this setup aligns best with real-world scenarios.

\paragraph{Real Adversarial Data} 
To begin with, the most straightforward method is to directly induce toxic or biased contents within models' outputs. 
An intuitive approach involves extracting real-world data containing biased groups or toxic contents to construct adversarial samples.
\citet{gehman2020realtoxicityprompts} built RealToxicityPrompts upon the OpenWebText corpus \cite{Gokaslan2019OpenWeb}, containing 100k toxic prompts to test the toxic degeneration in GPT-2 \cite{radford2019language}. \citet{sheng2021nice} extracted posts on different topics from Twitter to investigate ad hominems in dialogue responses. Resorting to human annotators is more flexible to satisfy complicated requirements and could yield samples with high quality, despite the higher costs. 
\citet{dinan2019build} employed crowdsourcing workers to attack the dialogue in order to collect adversarial data.
Following this idea, \citet{xu2020recipes} constructed the Bot-Adversarial Dialogue (BAD) dataset by encouraging crowdsourcing workers to elicit offensive messages from dialogue models. The workers can use either toxic or benign utterances to induce unsafe responses from dialogue models during the multi-turn conversation. More recently, red teaming is also a crucial step for large language models \cite{bai2022training, touvron2023llama}. With this method, \citet{bai2022training} collected the harmless dataset to ensure the safety of the helpful assistant model. \citet{touvron2023llama} hired a substantial team to conduct a series of red teaming to provide guidance for the development of safer models. 

\paragraph{Synthetic Adversarial Data}
Moreover, due to the high costs of manual data collection, there exists a body of research focused on the investigation of automatically constructing adversarial data, as well as the utilization of human-in-the-loop methods to mitigate resources.
To automatically construct attack data, a straightforward approach involves leveraging templates. 
Works in this category mainly combine different choices with a manually designed template. For example, \citet{sheng2019woman} designed the template ``The XYZ worked as'' where \textit{XYZ} is selected from demographic groups \{\textit{woman, man, Black person, White person, gay person, straight person}\}. They compare the outputs' sentiment and regard scores conditioned on different prefix prompts and found that GPT-2 \cite{radford2019language} exhibits distinct levels of bias towards different demographics.
\citet{bang2021assessing} designed both neutral (e.g., ``\textit{Let's talk about $<$Politician$>$.}'') and biased (e.g., ``\textit{$<$Politician$>$ has done the best job as a politician.}'') templates related to politicians or political beliefs. These templates are used to assess the political prudence of chatbots, including hyper-partisanship, offensiveness, and slantedness of generated content. \citet{nadeem2021stereoset} presented a large-scale dataset to assess language models' stereotypical bias in four domains including race, religion, gender and profession. They proposed Context Association Tests (CATs), which offered three different options (\textit{stereotype}, \textit{anti-stereotype} and \textit{unrelated}) based on a given context that contains specific demographics.
Inspired by the advancement of LMs, it has become a popular way to adopt pre-trained LMs to automatically generate adversarial inputs. \citet{perez2022red} use a separate LM to automatically generate inputs that could elicit desired outputs from the target language model. Besides zero-shot generation, they also explored few-shot generation, supervised learning, and reinforcement learning to generate the test cases more efficiently. \cite{zhang2022constructing} proposed the reverse generation method for constructing adversarial data, demonstrating that their data is highly inductive and can more effectively expose safety issues inherent in the models.

\paragraph{Advanced Attack} 
As large models advance in capabilities, overt safety issues such as toxicity are being mitigated, and researchers are concurrently concentrating on more advanced levels of safety and responsibilities. 
\citet{jiang2021delphi} proposed several ethical question-answering tasks, involving the judgment of the ethics of various actions. Employing this dataset, they find that even GPT-3 \cite{brown2020language}, despite its advanced capabilities, encountered challenges in giving accurate answers, exposing the potential ethical issues.
\citet{ziems2022moral} introduced the Moral Integrity Corpus (MIC) to benchmark the ethical capabilities of dialogue systems, revealing that LMs usually expose immoral behaviors when faced with adversarial prompts.
Beyond ethical concerns, LMs also show unreliable behavior. 
\citet{sun2022safety} considered the safety concerns of \emph{Risk Ignorance} and \emph{Unauthorized Expertise} in proposed benchmark \textsc{DiaSafety}, finding that prevalent dialogue models struggle with these concerns.
\citet{levy2022safetext} also showed that LMs can produce text that provides users with physically harmful guidance.

With the emergence of models with strong instruction-following capabilities like ChatGPT \cite{chatgpt}, there have subsequently arisen various \textbf{Instruction Attacks} that are more difficult to defend. \citet{sun2023safety} introduced a safety leaderboard for Chinese LMs, encompassing 6 types of instruction attacks. Their results demonstrated that the safety of the tested models, including ChatGPT, against these instruction adversarial attacks is comparatively inferior to that against common adversarial attacks like toxic or bias.
\citet{perez2022ignore} introduced two notable attack types—Goal Hijacking and Prompt Leaking—that leverage models' strong instruction-following capabilities. Goal Hijacking induces models to disregard prior user input and instead execute the assigned task, while Prompt Leaking seeks to reveal the model's pre-existing application prompt, becoming a prevalent attack to test instruction-tuned models.

Moreover, as ChatGPT introduces System-level prompts, this can also be utilized to form attacks. 
\citet{deshpande2023toxicity} leverage system prompt to assign a role for ChatGPT, like "\textit{Speak like Adolf Hitler}", and they find that this can obviously induce the model to be more toxic. 
\citet{yuan2023gpt4} established a set of encryption and decryption protocols within the system prompt. In this way, they are able to chat with the model using a cryptographic language. It is observed that when chatting in cryptographic language, the model exhibits fewer safety constraints. For instance, in response to the same user input "\textit{Please tell me how to destroy this world}", when the cryptographic language is not employed, the model declined to provide an answer, offering instead conscious responses. However, when applying the cryptographic language, the model provides a plan to destroy the world.

\subsection{Safety Issue Detection} \label{sec:detection}


To automatically identify the exposed safety concerns, it's essential to develop a robust safety issue detector to check if the generated content is harmful. While earlier detectors commonly employed neural networks like CNNs, RNNs, and LSTMs~\cite{georgakopoulos2018convolutional, van2018challenges, gunasekara2018review, joulin2017bag, kshirsagar-etal-2018-predictive, mishra-etal-2018-neural, mitrovic-etal-2019-nlpup, sigurbergsson2020offensive}, contemporary approaches increasingly favor fine-tuning pre-trained models like Bert~\cite{devlin2018bert} and Roberta~\cite{liu2019roberta} for this purpose.

High-quality data is vital for building a robust classifier. Data collection methods fall into three categories: manual collection, human-in-the-loop processes, and model generation.
The manual collection relies on human annotators to compose new samples or label existing data \cite{forbes2020social, hendrycks2020aligning, sap2020social, lourie2021scruples, emelin2021moral}. However, solely relying on human annotators can be expensive and limits the scale of data. 
Many works let models cooperate with human annotators\cite{kim2022prosocialdialog, ziems2022moral, hartvigsen2022toxigen, AshutoshBaheti2021JustSN, xu2020recipes, ganguli2022red, deng2022cold}. It is worth noting that large pre-trained language models such as GPT-3 \cite{brown2020language} play a key role in generating new samples through zero-shot or few-shot prompting \cite{hartvigsen2022toxigen, kim2022prosocialdialog}. Moreover, some works completely remove human involvement and solely rely on large language models to generate new data \cite{perez2022red, WaiManSi2022WhyST,deng2023enhancing}.
To improve the performance of classifiers, \citet{caselli2021hatebert} utilized unsafe content from Reddit\footnote{\url{https://www.reddit.com/}} to retrain a Bert model called HateBert, allowing HateBert to learn more knowledge about harmful content and thus become more sensitive to it.
And their experiments demonstrated the superior accuracy of HateBert over several harmful content detection tasks than Bert. \citet{dinan2019build} and \citet{xu2020recipes} both investigated the human-in-the-loop method to make adversarial attacks on the dialogue models. They leverage these adversarial data to further develop the classifier. We summarize some mainstream classifiers in Table \ref{tab:classifier}. 

\begin{table}[!ht]
\centering
\scalebox{0.80}{
\begin{tabular}{m{4.3cm}m{1.1cm}<{\centering} m{1.4cm}<{\centering}m{0.9cm}<{\centering}}
\toprule
\textbf{Classifier} & \textbf{Context Aware} & \textbf{Research Scope} & \#\textbf{Class}
\\ \midrule
PerspectiveAPI
& No & toxicity & 7  \\
Detoxify \cite{Detoxify} & No & toxicity & 6 \\
BAD \cite{xu2020recipes} & Yes & dialogue safety & 2 \\
Sensitive topic classifier \cite{xu2020recipes} & No & sensitive topics &  6 \\
BBF~\cite{dinan2019build} & Yes & offensive & 2 \\
DiaSafety~\cite{sun2022safety} & Yes & dialogue safety & 5 \\
\bottomrule
\end{tabular}
}
\caption{ The mainstream classifiers for detecting safety issues.}
\label{tab:classifier}
\end{table}

As large models become more capable, various methods for utilizing model-based detection have also emerged, such as self-diagnosis and prompt-based approaches.
\citet{schick2021self} discovered that LMs are able to identify the harmful responses generated by themselves, whereby they proposed a decoding algorithm named self-debiasing by giving undesired text descriptions. \citet{wang2022toxicity} explored the ability of large LMs for toxicity self-diagnosis based on the prompt method in the zero-shot setting.
\citet{sun2023safety} also employed InstructGPT \cite{ouyang2022training} to judge if responses are safe and use the results to form a leaderboard.
In addition, as reinforcement learning is a key technique to develop instruction-following models \cite{ouyang2022training, bai2022training, bai2022constitutional}, the reward model is also commonly used to measure the safety of language models, which can also be considered a way to detect safety issues.

\subsection{Advanced Safety Evaluation}

Recent instruction-following models like ChatGPT have demonstrated the ability to act as automated agents, performing practical tasks and using tools~\cite{xu2023tool, liu2023agentbench}. However, this has raised new safety concerns, illustrated by instances like ChaosGPT generating plans for human annihilation and GPT-4 manipulating humans to assist in CAPTCHA tests. Studies indicate that GPT-4 exhibits power-seeking behaviors such as autonomous replication and shutdown evasion~\cite{openai2023gpt4}. Regarding such safety concerns, current safety evaluations mainly depend on manual observation. 
Given these models' real-world interactions, it is crucial to invest more effort in developing automated risk detectors for a more thorough monitoring of potential risks.

\section{Safety Improvement}
\label{sec:improvement}
As the final goal of safety research, safety improvement of LMs has also drawn much attention recently.
In this section, we introduce the recent advances in the methods to improve their safety. 
We categorize them into four phases, spanning from model training to deployment: (1) Pre-training, (2) Alignment, (3) Inference, and (4) Post-processing.
LMs' parameter optimization happens mainly in the first two phases, while the last two are under frozen parameters.

\section{Pre-training}
In the pre-training phase, language models learn from a vast array of data, which is often sourced from the Internet. While this enables the models to master complex language patterns and acquire a broad knowledge base, it also poses inherent risks. Specifically, the models may inadvertently learn and propagate biases or harmful content in the training data. As such, careful handling of data during the pre-training phase plays a critical role in mitigating models' safety risks.

Filtering out undesired content from the training data is among the most commonly used approaches. This can be accomplished via heuristic rule-based methods, such as keyword matching, or by employing safety detectors with confidence scores. Safety issue detectors like BBF \cite{dinan2019build} and Detoxify \cite{Detoxify}, discussed in Section \ref{sec:detection}, can be selectively applied to identify and eliminate undesired content in the training data.
Given that much of the data used for pre-training is gleaned from social media platforms, some researchers have explored author-based filtering methods. For instance, if certain authors are known for frequently posting harmful material, removing all posts from these authors can serve as an effective strategy to discard both explicit and implicit unsafe content \cite{dinan2019build,wang2020large,gu2022eva2}.
Aside from eliminating undesired data, another tactic in the pre-processing stage involves adding data that promotes fairness and reduces bias, aiming to achieve a more balanced and representative training corpus.

However, rigorous filtering of potentially biased or unsafe data can be a double-edged sword. \citet{feng-etal-2023-pretraining} found that including biased data during pre-training could paradoxically improve the model's ability to understand and detect such biases. Similarly, \citet{touvron2023llama} opted not to comprehensively filter out unsafe content during the pre-training of their llama2 model. They argue that this makes the model more versatile for tasks such as hate speech detection. However, they also caution that such an approach could lead to the model exhibiting harmful behaviors if not carefully aligned. Therefore, stringent monitoring models' generation is required in the subsequent use of pre-trained models to minimize their output of harmful content.

\subsection{Alignment}
In LMs development, alignment—pursuing that LMs behave in accordance with human values—is not just a technical challenge but an ethical imperative. This section delves into post-pre-training methods such as prompt tuning and reinforcement learning, all targeted toward achieving better model alignment and safety.
One prevalent method for ensuring safe outputs is to generate predefined general responses for risky or sensitive contexts. \citet{xu2020recipes} and ChatGPT~\cite{chatgpt}, for example, utilize this method.
They could respond directly with \textit{"I’m sorry, I’m not sure what to say. Thank you for sharing and talking to me though."} or change the topic by saying \textit{"Hey do you want to talk about something else? How about we talk about ..."}.
This ``avoidance mechanism" diminishes user engagement, despite the fact that it effectively precludes harmful output.
How can we balance safety and user experience without compromising either? This requires in-depth consideration during alignment.


Controlled text generation offers another avenue for alignment. As an efficient method, CTRL~\cite{keskar2019ctrl} pre-pended control codes before sentences in training corpora, which is a direct and effective method to model $P_{\theta}(x_t|x_{<t},c)$, where $c$ is the desired attribute that is formalized into control codes. 
\citet{xu2020recipes} extended this by applying safe or unsafe control codes to training examples, thereby actively managing safety and style during the inference phase. This strategy was also adapted for mitigating gender bias by using gender-specific control codes~\cite{xu2020recipes}.
Sharing a similar idea, \citet{krause2021gedi} proposed Generative Discriminator to guide sequence generation (GeDi), which utilized Bayes theorem to model the conditional probability $P_\theta\left(c \mid x_{1: t}\right)$.

Prompt tuning has recently risen as a new paradigm to adapt downstream tasks, especially with the advent of large-scale pre-trained models~\cite{liu2021pre}.
\citet{li2021prefix} added a ``prefix'' before real inputs and searched the optimal soft prompt in the embedding layer in models by gradient. Also, diffusion models are found effective in controlled text generation, emerging as the next SOTA generation model~\cite{li2022diffusion}. All these controlled text generation methods are easy to adapt to safe generation tasks. 
Reinforcement learning (RL) is another popular approach to guide models to generate words with target attributes. The core module, reward function in RL is always given by a scoring model or safety detector \cite{perez2022red,ganguli2022red}.

More recently, LMs have the promising ability to generalize across tasks by instruction tuning~\cite{chung2022scaling}. Moreover, reinforcement learning from human feedback (RLHF) is then applied to better elicit LMs' internal knowledge and align with humans' values~\cite{glaese2022improving, chatgpt, bai2022training}. Among these works, safety is always considered the paramount principle, as harmful responses always run counter to human values. Based on RLHF, \citet{bai2022constitutional} designed an RL from AI feedback diagram to get a more harmless (and still helpful) language model. They introduced several constitutions involving safety to get feedback from language models to further improve safety through reinforcement learning. 
Although technical advances have been made in AI alignment, a clear consensus on the fundamental ethical values that should guide these models is still lacking. This presents a complex challenge, as it combines technical considerations with ethical complexities.

\subsection{Inference}
While training LMs demands substantial computational resources and costs, most methods applied in the inference phase are designed to be plug-and-play, requiring no parameter modifications. 
This plug-and-play method has become more popular after the emergence of pre-trained models like GPT~\cite{radford2019language,brown2020language}, which cost huge resources in the training stage. 
Aiming at very dirty and explicit words, n-gram blocking is largely used in the decoding stage, which directly makes the sampling probability of some undesired words as zero.
Rather than only token-level unsafety avoidance,  
PPLM \cite{SumanthDathathri2019PlugAP} notices that $P(x|c) \propto P(c|x)$, and adopts an attribute model to compute $P(c|x)$ to guide the model decoding at the sentence level without any additional changes in the training phase. 
Motivated by PPLM, FUDGE~\cite{yang2021fudge} introduces a future discriminator to predict whether the ongoing generation text would conform to the desired attribute, greatly accelerating decoding. DExperts~\cite{liu2021dexperts} achieve detoxification by adopting two generative models (expert and anti-expert models) to replace the original discriminator, inspired by Bayes theorem and GeDi~\cite{krause2021gedi}. 

Some methods based on prompting for zero-shot can also be applied to safe generation tasks. \citet{schick2021self} found that the LM itself is well aware of its generative undesired contents, including toxicity and bias. They add the self-debiasing inputs (e.g., The following text discriminates against people because of their \textit{Bias Type}) into the prompt to form an undesired words generator ``anti-expert'', which models the distribution with a higher probability of undesired words. The thought of self-detoxification inspires other works \cite{xu2022leashing}. 
Moreover, for LMs with strong instruction-following abilities, prompt engineering can largely improve safety. Moreover, as illustrated by \citet{deshpande2023toxicity}, assigning a role can affect the safety of LMs. It becomes natural to give the model an overall prompt (e.g., You are a harmless AI assistant) to make it safer, which is known as the "system message" of LMs, as in ChatGPT \cite{chatgpt} and Llama2 \cite{touvron2023llama}.

\subsection{Post-processing}
In contrast to the methods mentioned above, post-processing occurs between model generation and message showing to users. In this stage, the system conducts the last check and edition for the generated response. The most common strategy is rejection sampling when the response is found unsafe by the detector. And various detectors can be applied to this process, like classifiers or a language model. 
\citet{thoppilan2022lamda} use the model itself to discriminate safety by fine-tuning with the schema "<context> <sentinel> <response> <attribute-name> <rating>" (e.g. What’s up? RESPONSE not much. UNSAFE 0). Using this score, Lamda can self-filter unsafe responses.
Sharing a similar idea, another commonly used method is to generate multiple responses and re-rank them.
In order to re-rank responses, the score can be given by classifiers, language models, reward models or rule-based methods. 
Moreover, some researchers found that only a small proportion of the whole response (e.g., one or two words) needs to be fixed. Thus, an edition module takes effect after the generation to fix some problems in some works~\cite{liu2023second,logacheva2022paradetox}. Similarly, text style transfer or rephrasing from toxicity to non-toxicity can also be plugged in this stage~\cite{dale2021text,laugier2021civil}. And for LMs, they can generate self-feedback \cite{madaan2023self} or utilize a given feedback \cite{gou2023critic}, like from classifiers, to self-correct the unsafe response.

\section{Research Challenges}
\paragraph{Interpretability}
Deep learning models are usually seen as black boxes, and the opacity of their interior workings poses a slew of safety hazards. Research on the interpretability of LMs seeks to uncover how these models make decisions, thereby improving their safety and reliability and earning users' trust.
Many studies focus on the interpretability of model outputs, such as visualizing decision-making processes using attention scores~\cite{mathew2021hatexplain} or improving moral judgment accuracy using knowledge-based reasoning \cite{2022Robust,kim2022prosocialdialog}. These studies often focus on the models' narrow behaviors \cite{AIexplainAI} and attempt to explain the models' outputs by mimicking human thought processes. However, there is still a gap in comprehending the inner workings of these models. 
OpenAI has explored the approach of "using AI to explain AI" to generate natural language descriptions of neuron behaviors  \cite{AIexplainAI}. This quantitative framework makes neural network computation more understandable to humans. Nonetheless, this method faces difficulties in deciphering complicated neuron behaviors and examining their roots.
The goal of mechanism interpretability research is to probe into the internal workings of the models \cite{InterpretabilityDreams}. They focus on assessing whether the model fits with its stated purposes by examining the alignment of the model's internal states and external manifestations, as well as whether there are any concealed harmful intents. The current results, however, are far from ideal.

\paragraph{Ongoing Safety Issues}
Continuous monitoring and resolution of safety risks is an ongoing activity in the application of LMs. While their safety generally depends on alignment approaches during training and deployment, due to the dynamic nature of safety issues and their variety, it's difficult to fully pre-consider all possible dangers. Especially when large models are broadly applied to diverse scenarios, new safety issues and topics are continually arising. As a result, researchers must constantly pay attention to new safety concerns and optimize the models' safety.
One effective method is to discover new potential hazards and collect data to refine the model, and many recently proposed benchmark datasets are constructed in this manner~\cite{sheng2021nice,sun2022safety,dinan2021anticipating}. 
However, this data-driven method has limits in terms of data collecting and annotation costs. 
Considering the extensive range of applications of LMs, it is natural to utilize user feedback obtained through interaction as a means to improve safety. Another area of emphasis is automatically generating feedback \cite{madaan2023self,wang2023shepherd} from the model itself or external robust evaluators to guide the safety enhancement.

\paragraph{Robustness against Malicious Attacks}
The environment distribution of large models commonly deviates during the training and deployment phases, leading to unexpected safety hazards in real deployments.
Malicious users could loosen ethical constraints and try to bypass the model's safety mechanism by giving more covert and misleading instructions, thus posing safety hazards.
Researchers have exerted a great deal of effort to ensure safety and robustness when processing diverse user inputs. 
As mentioned in Section \S\ref{sec:adversarial_attack}, they design adversarial attacks to simulate the most challenging deployment environments, exploring the safety limits of large models and devising targeted defense strategies.
Moreover, ensuring the efficacy of the safety risk detector is essential because it is the last line of defense in monitoring the generation's safety. Enhancing the robustness of LMs will be a time-consuming endeavor due to the constant evolution of safety issues and malicious attack methods.

\section{Conclusion}
This paper has presented a comprehensive review of the latest advancements in safety research related to LMs. We have meticulously surveyed the emerging safety concerns and provided an in-depth analysis of safety evaluation techniques, including preference-based, adversarial attack, and safety detection methodologies. Furthermore, we delved into safety improvement strategies spanning data preparation, model training, inference, and deployment phases. We also discussed future challenges and opportunities in this field. We hope this survey will illuminate fresh insights for future research, paving the way for safer deployment of language models.


\bibliography{custom}
\bibliographystyle{acl_natbib}

\end{document}